\documentclass{article}

\usepackage{arxiv}

\usepackage[T1]{fontenc}    % use 8-bit T1 fonts
\usepackage{hyperref}       % hyperlinks
\usepackage{url}            % simple URL typesetting
\usepackage{booktabs}       % professional-quality tables
\usepackage{multirow}       % multi-row cells in tables
\usepackage{amsmath}        % math extensions
\usepackage{amsfonts}       % blackboard math symbols
\usepackage{nicefrac}       % compact symbols for 1/2, etc.
\usepackage{microtype}      % microtypography
\usepackage{lipsum}
\usepackage{fancyhdr}       % header
\usepackage{graphicx}       % graphics
\graphicspath{{media/}}     % organize your images and other figures under media/ folder

\title{Demystifying KAN for Vision Tasks: The RepKAN Approach}

\author{
  Minjong Cheon \\
  Sejong University \\
  Computer Science and Engineering \\
  Seoul \\
  \texttt{jmj2316@sejong.ac.kr} \\
}

\begin{document}
\maketitle

\begin{abstract}
Remote sensing image classification is essential for Earth observation, yet standard CNNs and Transformers often function as uninterpretable black-boxes. We propose RepKAN, a novel architecture that integrates the structural efficiency of CNNs with the non-linear representational power of KANs. By utilizing a dual-path design---Spatial Linear and Spectral Non-linear---RepKAN enables the autonomous discovery of class-specific spectral fingerprints and physical interaction manifolds. Experimental results on the EuroSAT and NWPU-RESISC45 datasets demonstrate that RepKAN provides explicit physically interpretable reasoning while outperforming state-of-the-art models. These findings indicate that RepKAN holds significant potential to serve as the backbone for future interpretable visual foundation models.
\end{abstract}

\section{Introduction}
The advancement of remote sensing technology has significantly heightened interest in high-resolution Earth observation. Remote sensing image classification, serving as the bedrock of intelligent interpretation, is a crucial element for subsequent downstream tasks such as land mapping, environmental monitoring, and urban planning \cite{li2024remote}. Nonetheless, the inherent complexity and spectral diversity of remote sensing scenarios, coupled with the variable spatio-temporal resolution, present substantial challenges to automated and reliable classification. Recent overviews and modern sequence-model approaches further highlight both the progress and the remaining challenges in this domain \cite{cheon2025mambaoutrs,chen2024rsmamba}.

Researchers have been diligently working towards alleviating these challenges by enhancing model representational power. Early methodologies predominantly focused on manual feature engineering, with hand-crafted descriptors such as SIFT and LBP and traditional spectral indices such as NDVI and NDWI~\cite{lowe2004sift,ojala2002lbp}. In recent years, deep learning has revolutionized this paradigm by enabling end-to-end feature learning from raw data~\cite{krizhevsky2012alexnet}. In terms of network architecture, representative families include CNNs for local spatial abstraction and attention-based networks such as ViT and Swin for long-range dependency modeling~\cite{dosovitskiy2021vit,liu2021swin}. Despite strong performance, these models often remain difficult to interpret as transparent decision systems, which motivates explainability-focused analyses in remote sensing~\cite{selvaraju2017gradcam,li2024remote}.

To a certain extent, the classification accuracy of deep learning heavily depends on the model's ability to effectively handle complex spectral-spatial interactions~\cite{li2024remote}. While post-hoc Explainable AI (XAI) techniques such as Grad-CAM provide spatial saliency maps, they often fail to explain the non-linear spectral dynamics that are essential for physical interpretation in remote sensing~\cite{selvaraju2017gradcam}. KAN, which replaces static activation functions with learnable 1D splines, presents an attractive direction for addressing this transparency gap~\cite{liu2024kan}. However, its original formulation still poses significant challenges in vision tasks; specifically, flattening image inputs in vanilla KAN discards the local spatial context that is essential for land-cover structure analysis~\cite{cheon2024demonstrating}\cite{liu2024kan}\cite{liu2024kanb}.

In this paper, we introduce RepKAN, an efficient and interpretable hybrid module for multispectral remote sensing image classification. RepKAN integrates spatial convolutions with 1D KAN splines to preserve local context while modeling non-linear spectral interactions. The module can be plugged into standard CNN backbones, where local convolutions capture spatial features and learnable B-splines model band interactions. This design improves performance and enables data-driven spectral index discovery.

The main contributions of this study are summarized as follows:

\begin{itemize}
  \item \textbf{Structural Hybridization for Vision-KAN.} We propose RepKAN, a plug-and-play module that effectively adapts KANs for computer vision. By integrating spatial convolutions with spectral 1D splines, RepKAN overcomes the spatial information loss inherent in vanilla KANs while consistently enhancing classification performance in multispectral tasks.

  \item \textbf{Intrinsic Interpretation of Spectral Dynamics.} We introduce an analytical framework to visualize the internal dynamics of spectral-spatial interactions. Unlike post-hoc saliency maps, RepKAN provides intrinsic transparency by mapping band-wise energy distributions and non-linear interaction trajectories, offering a granular understanding of the model's decision-making process.
  
  \item \textbf{Symbolic Synthesis of Physics-aware Equations.} We demonstrate the model's capacity to autonomously discover mathematical formulations. By performing symbolic regression on learned expert filters, we extract explicit non-linear equations that rediscover and refine classical physical indices, offering a human-readable, data-driven bridge to traditional remote sensing.
\end{itemize}

\section{Methodology}
\label{sec:method}

Leveraging the inherent characteristics of KAN, RepKAN is proficient in effectively capturing non-linear spectral interactions while preserving spatial structural information. This section begins with the preliminaries of KAN, followed by an overview of the RepKAN architecture. Subsequently, we explore the RepKANLayer in depth, focusing on its hybrid spatial-spectral modeling and structural reparameterization.

\subsection{Preliminaries: KANs}
The KAN is derived from the Kolmogorov-Arnold representation theorem, which posits that any multivariate continuous function can be represented as a finite sum of univariate continuous functions~\cite{liu2024kan}. Unlike traditional Multi-Layer Perceptrons (MLPs) that apply static activation functions on nodes, KAN applies learnable activation functions on the edges~\cite{liu2024kan}. The univariate function $\phi(x)$ on each edge is formulated as:
\begin{equation}
    \phi(x) = w \cdot (b(x) + s(x))
\end{equation}
where $b(x) = \text{silu}(x)$ denotes the base function, and $s(x)$ represents the spline function. The spline is constructed as a linear combination of B-spline basis functions $B_i(x)$:
\begin{equation}
    s(x) = \sum_{i} c_i B_i(x)
\end{equation}
where $c_i$ are learnable coefficients. This architecture allows KAN to achieve high approximation accuracy and intrinsic interpretability through learnable spline trajectories~\cite{liu2024kan}\cite{cheon2024kolmogorov}\cite{cheon2024combining}.

\subsection{RepKAN Architecture}
RepKAN transforms multispectral images into hierarchical feature representations through a sequence of hybrid blocks. Given a multispectral image $\mathcal{I} \in \mathbb{R}^{H \times W \times C}$ (where $C=13$ for the EuroSAT benchmark), we first map local patches into pixel-wise feature embeddings. The hierarchical modeling process is delineated as follows:
\begin{equation}
    \mathbf{T}_i = \Phi_{\text{RepKAN}_i}(\mathbf{T}_{i-1}) + \mathbf{T}_{i-1}
\end{equation}
where $i$ signifies the $i$-th stage and $\mathbf{T}_i$ represents the output feature map. Following the final stage, a Global Average Pooling (GAP) operation is applied to derive the semantic representation $\hat{y}$ for category prediction:
\begin{equation}
    \hat{y} = \Phi_{\text{proj}}(\Phi_{\text{GAP}}(\mathbf{T}_N))
\end{equation}
where $\Phi_{\text{proj}}$ projects the latent features to the number of target classes. For efficient inference, the spatial convolutional branches are structurally re-parameterized into a single branch in the spirit of RepVGG~\cite{ding2021repvgg}.

\subsection{RepKANLayer: Hybrid Spatial-Spectral Modeling}
The vanilla KAN encounters difficulties in modeling spatial positional relationships due to its inherent 1-D vector processing. To alleviate this, we introduce the \textbf{RepKANLayer}, which operates on 2-D feature maps using a dual-path mechanism. The output of the $i$-th RepKANLayer is formulated as:
\begin{equation}
    \mathbf{Y} = \mathcal{F}_{\text{spatial}}(\mathbf{X}) \oplus \mathcal{F}_{\text{spectral}}(\mathbf{X})
\end{equation}
where $\mathcal{F}_{\text{spatial}}$ and $\mathcal{F}_{\text{spectral}}$ denote the spatial convolutional path and the spectral spline path, respectively.

\subsubsection{Spatial Path}
To capture local structural features and spatial context, we utilize a multi-branch convolutional structure:
\begin{equation}
    \mathcal{F}_{\text{spatial}}(\mathbf{X}) = \text{BN}(\text{Conv}_{1\times1}(\mathbf{X})) + \text{BN}(\text{Conv}_{3\times3}(\mathbf{X}))
\end{equation}
This path ensures that the model retains the robust spatial abstraction capabilities of traditional CNNs.

\subsubsection{Spectral Path}
The spectral interactions are modeled using 1-D B-splines applied along the channel dimension, enabling the discovery of data-driven spectral indices. For each output channel $o$, the response is calculated as:
\begin{equation}
    \mathcal{F}_{\text{spectral}}(\mathbf{X})_{o,h,w} = \sum_{c=1}^{C_{\text{in}}} \phi_{o,c}(\mathbf{X}_{c,h,w})
\end{equation}
This path functions as a symbolically explicit representation learner, where the learned splines $\phi_{o,c}$ can be extracted to formulate explicit mathematical equations for diverse land-cover materials.

\begin{figure}[t]
  \centering
  \includegraphics[width=\linewidth]{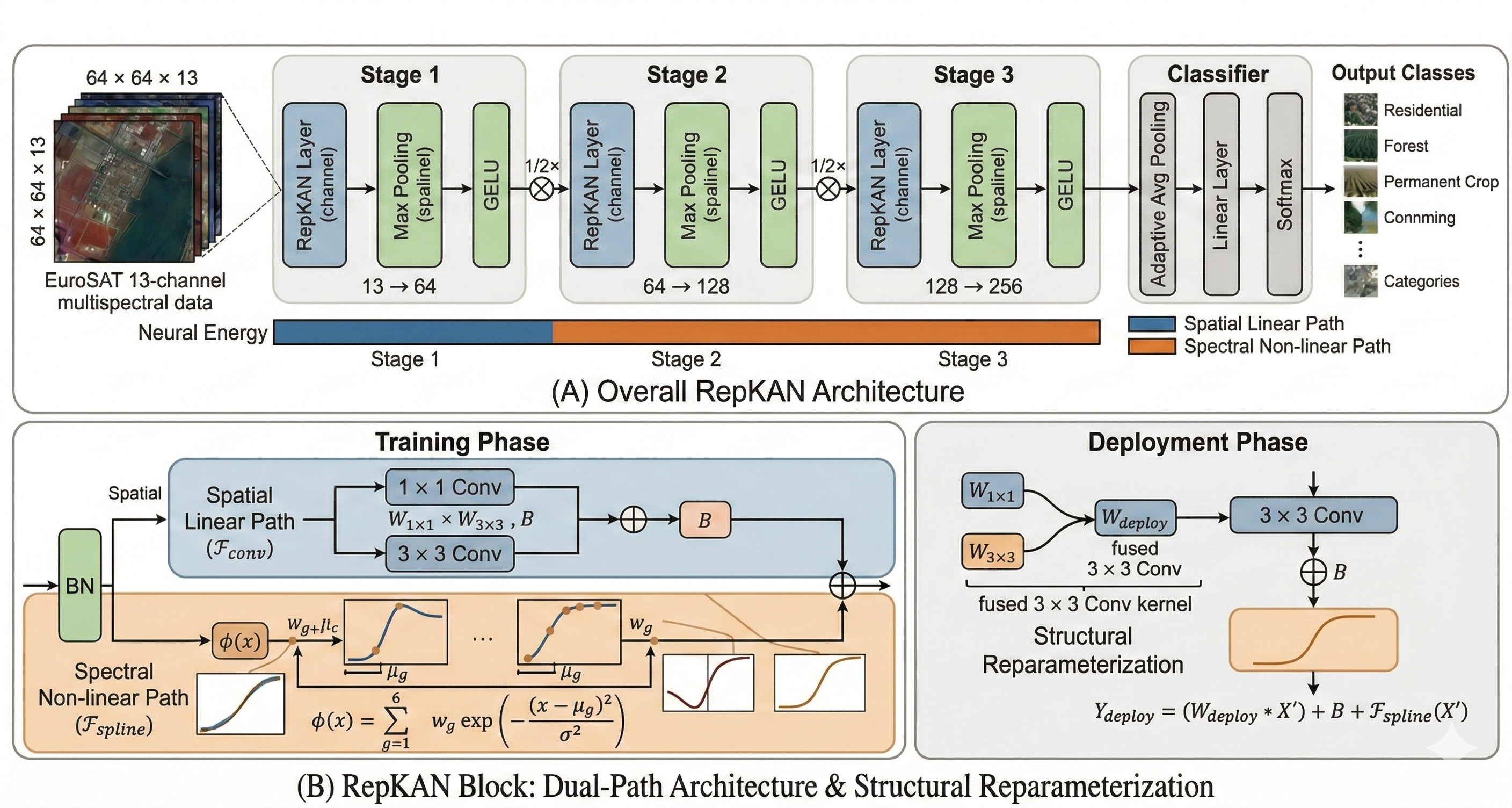}
  \caption{Overview of the RepKAN architecture and the proposed RepKAN block.(A) Hierarchical structure of RepKAN for 13-channel multispectral image classification. The network progressively abstracts features across three stages, with the neural energy bar illustrating the transition from spatial to spectral focus. (B) Detailed design of the RepKAN block. During training, it utilizes a dual-path mechanism: a Spatial Linear Path for local spatial context and a Spectral Non-linear Path ($\mathcal{F}_{spline}$) featuring learnable activation functions $\phi(x)$ to model channel-wise band interactions. For efficient deployment, the spatial branches are mathematically fused into a single $3\times3$ convolution ($\mathcal{W}_{deploy}$) via structural reparameterization, computing the final output as $Y_{deploy} = (\mathcal{W}_{deploy} * X') + B + \mathcal{F}_{spline}(X')$.}
\end{figure}

\section{Experimental Results and Analyses}

In this section, we conduct a series of experiments to evaluate the performance and interpretability of the proposed RepKAN module. We first describe the datasets used for evaluation and the specific implementation details of our training pipeline. Subsequently, we provide a comprehensive analysis of the experimental results.

\subsection{Dataset Description}
We conducted experiments on two benchmark datasets to evaluate the efficacy of RepKAN across diverse remote sensing scenarios, : EuroSAT (13-channel multispectral) and NWPU-RESISC45 (RESISC45)~\cite{helber2019eurosat,cheng2017resisc45}.

\begin{itemize}
    \item \textbf{EuroSAT~\cite{helber2019eurosat}:} This dataset consists of Sentinel-2 satellite images covering 10 distinct land-use categories with an aggregate of 27,000 images. Unlike standard RGB datasets, we utilize the 13-channel multispectral data to fully exploit the spectral modeling capabilities of RepKAN. Each image has a $64 \times 64$ pixel resolution. We designated 80\% of the images for training and 20\% for validation.
        
    \item \textbf{RESISC45~\cite{cheng2017resisc45}:} The RESISC45 comprises 31,500 remote sensing images obtained from Google Earth, segregated into 45 scene categories. Each category contains 700 RGB images with $256 \times 256$ pixel resolution. The spatial resolution fluctuates between approximately 30m to 0.2m per pixel. We allocated 70\% of the images from each category for training.
\end{itemize}

\subsection{Implementation Details}
To enhance the model's robustness, we implement data augmentation techniques including random cropping, horizontal flipping, and photometric distortion. For the RepKAN module, we evaluate three grid size configurations: $G \in \{3, 5, 7\}$. 

For supervised training, we employed the cross-entropy loss function and utilize the AdamW optimizer with an initial learning rate of $5 \times 10^{-4}$ and a weight decay of $0.05$. The learning rate is decayed using a cosine annealing scheduler with a linear warmup phase. Considering the efficient convergence of the KAN-based architecture, the training process spans a total of \textbf{50 epochs} with a batch size of 1024. All experiments were implemented using the PyTorch framework and executed on one NVIDIA A100 GPU. We employ Precision (P), Recall (R), F1-score (F1), and Overall Accuracy (OA) as the primary performance metrics to provide a multi-faceted evaluation of the model.

\subsection{Performance Comparison and Analysis}

We compared RepKAN with baseline CNN models across multiple datasets. Accuracy, Precision, Recall, and F1-score are summarized below.

\subsubsection{Results on EuroSAT (13-Channel)}
Table \ref{tab:eurosat_results} reports results on the 13-channel EuroSAT dataset. RepKAN outperforms the baseline CNN across all metrics, and RepKAN\_Grid3 achieves the best overall accuracy (\textbf{0.9878}).

Interestingly, we observed that increasing the grid size beyond 3 (Grid 5 and Grid 7) led to a slight marginal degradation in performance (0.9859 and 0.9856, respectively). This results aligns with the previous research, which also proved that lower grid number yielded better performance in image classification tasks \cite{joo2025bridging}.

\begin{table}[ht]
    \centering
    \caption{Final Evaluation of RepKAN on EuroSAT 13-channel Multispectral Dataset.}
    \label{tab:eurosat_results}
    \begin{tabular}{l|cccc}
        \hline
        \textbf{Model Name} & \textbf{Accuracy} & \textbf{Precision} & \textbf{Recall} & \textbf{F1 Score} \\
        \hline
        Baseline\_CNN\_13Ch & 0.9841 & 0.9830 & 0.9837 & 0.9833 \\
        \hline
        \textbf{RepKAN\_Grid3} & \textbf{0.9878} & \textbf{0.9872} & \textbf{0.9872} & \textbf{0.9871} \\
        RepKAN\_Grid5 & 0.9859 & 0.9854 & 0.9852 & 0.9853 \\
        RepKAN\_Grid7 & 0.9856 & 0.9846 & 0.9848 & 0.9847 \\
        \hline
    \end{tabular}
\end{table}

\subsubsection{Results on NWPU-RESISC45}
The results for the NWPU-RESISC45 dataset are summarized in Table \ref{tab:resisc_results}. The integration of the RepKAN module significantly enhances the baseline's performance on this complex aerial scene dataset. Specifically, RepKAN\_NWPU\_Grid3 achieved an accuracy of \textbf{0.7917}, representing an improvement of approximately \textbf{5.36\%} over the Baseline\_CNN\_NWPU (0.7381). This performance gain across all metrics (Precision, Recall, and F1-score) demonstrates the robust generalization capability of RepKAN in capturing high-level semantic features in diverse remote sensing scenarios.

\begin{table}[ht]
    \centering
    \caption{Final Evaluation of RepKAN on NWPU-RESISC45 Dataset.}
    \label{tab:resisc_results}
    \begin{tabular}{l|cccc}
        \hline
        \textbf{Model Name} & \textbf{Accuracy} & \textbf{Precision} & \textbf{Recall} & \textbf{F1 Score} \\
        \hline
        Baseline\_CNN & 0.7381 & 0.7423 & 0.7332 & 0.7309 \\
        \hline
        \textbf{RepKAN\_Grid3} & \textbf{0.7917} & \textbf{0.7962} & \textbf{0.7883} & \textbf{0.7889} \\
        \hline
    \end{tabular}
\end{table}

\subsection{Interpretation of Neural Energy and Autonomous Spectral Index Discovery}
\label{sec:interpretation}

\begin{figure*}[htbp]
    \centering
    \includegraphics[width=\textwidth]{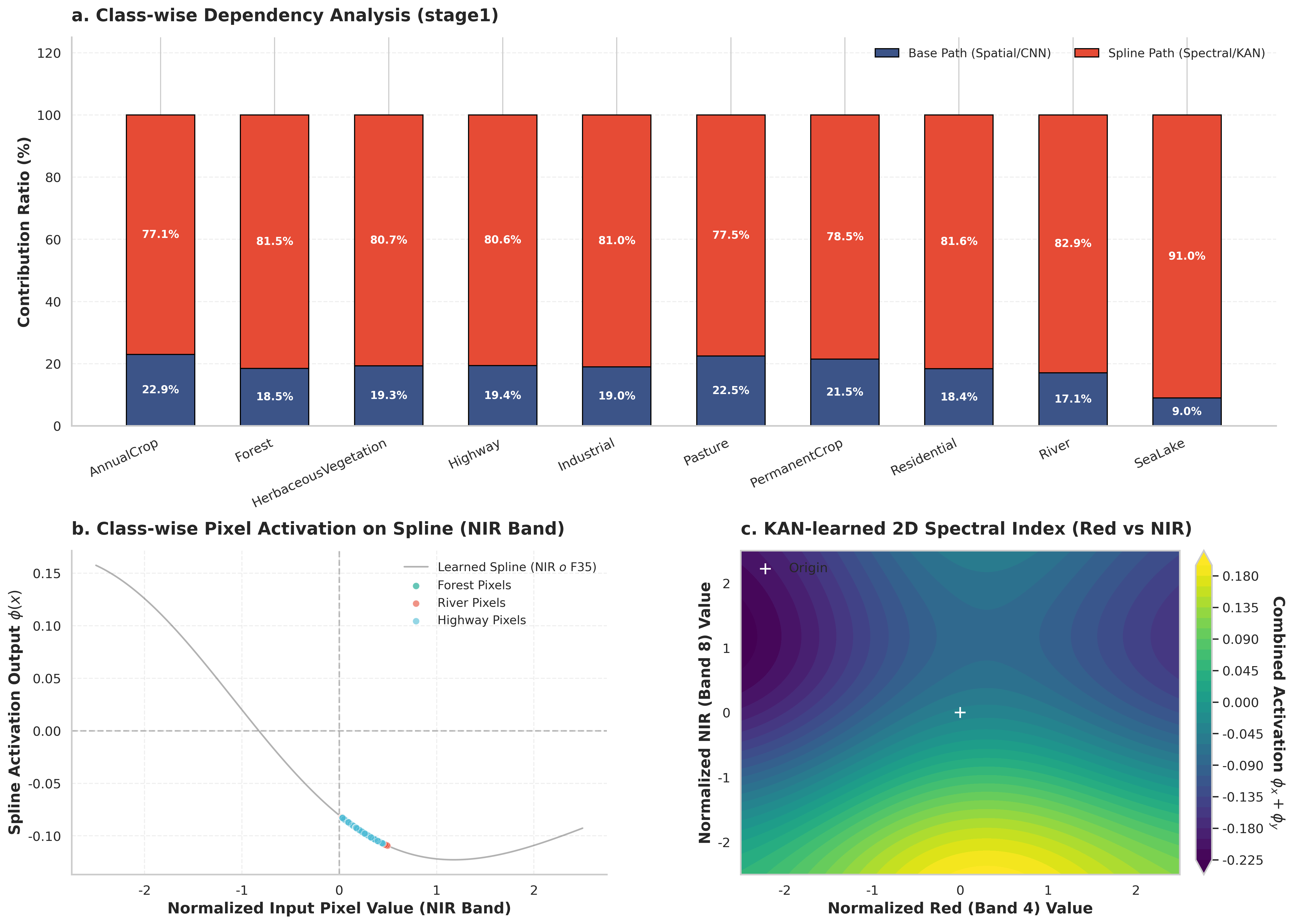}
    \caption{Comprehensive analysis of spectral indices in RepKAN. (\textbf{a}) Class-wise dependency analysis at Stage 1, showing the contribution ratio between the Base Path (Spatial/CNN) and the Spline Path (Spectral/KAN). The model exhibits a high dependency ($>77.1\%$) on non-linear spectral interactions for all categories, especially for \textit{SeaLake} ($91.0\%$). (\textbf{b}) 1D spline activation curve for the NIR band with overlaid pixel distributions of \textit{Forest}, \textit{River}, and \textit{Highway}. The learned gray spline acts as a non-linear projector that differentiates land-cover types based on their physical NIR reflectance. (\textbf{c}) KAN-learned 2D spectral interaction landscape between Red and NIR bands. The contour map represents a self-generated spectral-spatial index, demonstrating the model's ability to discover optimal band combinations for land-cover classification.}
    \label{fig:xai_consolidated}
\end{figure*}

To reveal the decision-making logic of RepKAN, we carried out an interpretability profiling using the Stage 1 outputs, as summarized in Fig.~\ref{fig:xai_consolidated}. 

First, we analyzed the energy contribution ratios across all land-cover categories to understand the model's feature dependency. As shown in Fig.~\ref{fig:xai_consolidated}(a), RepKAN exhibits a dominant reliance on the Spectral Non-linear Path (Spline Path) at the initial stage, with contribution ratios exceeding 77.1\% for all classes. Notably, the \textit{SeaLake} category demonstrates the highest spectral dependency at 91.0\%, which physically aligns with the NIR absorption and spatial homogeneity characteristic of water bodies, as the absence of distinctive textures often facilitates a more streamlined symbolic representation \cite{unger2007introductory}.

In Fig.~\ref{fig:xai_consolidated}(b), we projected the pixel-wise distributions of Forest, River, and Highway onto the learned 1D spline activation curve for the Near-Infrared (NIR) band. The model autonomously learns a non-linear mapping function that projects distinct physical materials onto non-overlapping segments of the activation manifold. Specifically, Forest pixels—characterized by the high NIR reflectance typical of healthy vegetation—are clustered in the high-intensity input region ($x > 0.5$) and mapped to a distinct activation response. This effectively demonstrates that the spline function serves as an interpretable spectral discriminator that isolates class-specific radiometric signatures.

Finally, Fig.~\ref{fig:xai_consolidated}(c) visualizes the 2D spectral interaction landscape between the Red (Band 4) and NIR (Band 8) channels. The synthesized activation surface ($\phi_x + \phi_y$) represents an autonomously discovered spectral index. The model formulates a complex, non-linear interaction manifold without human-defined priors, maximizing its representational power in the Red-NIR feature space to enhance classification robustness across diverse land-cover scenarios.

\subsection{Autonomous Discovery of Class-Specific Spectral Fingerprints}

\begin{figure*}[htbp]
    \centering
    \includegraphics[width=\textwidth]{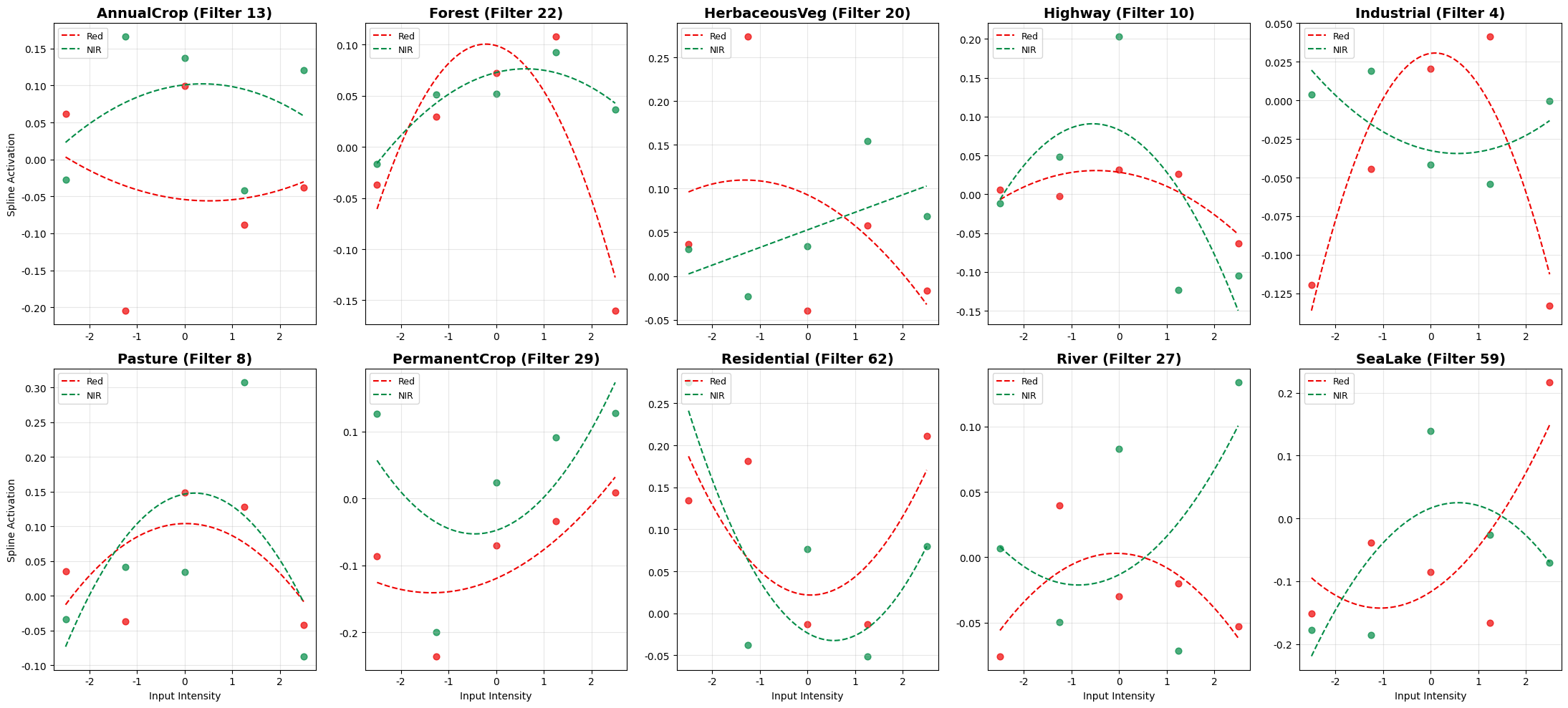}
    \caption{Interpretability analysis via learned spline activation functions for Red and NIR spectral bands. Each subplot displays the non-linear mapping learned by the model's spline layers (e.g., in a KAN-based architecture) for specific filters corresponding to various land-cover classes. The dashed lines represent the continuous learned activation functions, while the dots indicate the actual activation values across the input intensity range. These visualizations reveal how the model autonomously adapts its activation shapes---ranging from parabolic to sigmoidal---to capture unique spectral signatures, providing transparency into the decision-making process for remote sensing classification.}
    \label{fig:spectral_fingerprints}
\end{figure*}

To further evaluate the interpretability of the proposed model, we visualize the learned spline activation functions for the Red and NIR bands across different land-cover categories (Fig.~\ref{fig:spectral_fingerprints}). The expert filters for each class were identified through a statistical sensitivity analysis, selecting the neurons that exhibited the highest mean activation for their respective categories across the validation set. Unlike traditional deep learning models that rely on fixed activation functions, the KAN-based architecture allows each neuron to learn a specific functional mapping tailored to the input features \cite{liu2024kan}.

The diversity in the shapes of these learned splines provides a direct window into the model's ``reasoning.'' For instance, in the Forest (Filter 22) and PermanentCrop (Filter 29) classes, the model learns distinct non-linear responses for Red and NIR intensities, effectively mimicking the complex vegetation indices (e.g., NDVI) used in traditional remote sensing. Conversely, for Industrial (Filter 4) and SeaLake (Filter 59), the activation curves exhibit significantly different trends, reflecting the unique reflectance properties of man-made structures and water bodies.

By observing these transformations, we confirm that the model captures the underlying physical characteristics of the spectral data.

\begin{table*}[t]
\centering
\caption{Ablation Study on the Degree of Symbolic Discovery and the Discovered Cubic Equations for RepKAN Expert Filters across EuroSAT Categories.}
\label{tab:symbolic_ablation}
\resizebox{\textwidth}{!}{
\begin{tabular}{llccc|l}
\toprule
\multirow{2}{*}{\textbf{Class}} & \multirow{2}{*}{\textbf{Band}} & \multicolumn{3}{c|}{\textbf{$R^2$ Score by Polynomial Degree}} & \multirow{2}{*}{\textbf{Discovered Cubic Equation ($\phi(x)$)}} \\
\cmidrule{3-5}
& & \textbf{1st (Linear)} & \textbf{2nd (Quadratic)} & \textbf{3rd (Cubic)} & \\
\midrule
\multirow{2}{*}{AnnualCrop} & Red & 0.074 & 0.121 & 0.518 & $0.0173x^3 + 0.0075x^2 - 0.1056x - 0.0072$ \\
                            & NIR & 0.714 & 0.835 & 0.837 & $0.0020x^3 - 0.0214x^2 + 0.0659x + 0.0543$ \\
\midrule
\multirow{2}{*}{Forest}     & Red & 0.148 & 0.657 & 0.660 & $0.0021x^3 - 0.0350x^2 - 0.0390x + 0.1108$ \\
                            & NIR & 0.436 & 0.443 & 0.874 & $-0.0075x^3 + 0.0012x^2 + 0.0256x - 0.0171$ \\
\midrule
\multirow{2}{*}{HerbaceousVeg} & Red & 0.023 & 0.094 & \textbf{0.956} & $0.0180x^3 + 0.0066x^2 - 0.1011x + 0.0593$ \\
                               & NIR & 0.039 & 0.099 & \textbf{0.782} & $0.0380x^3 - 0.0144x^2 - 0.1851x + 0.0412$ \\
\midrule
\multirow{2}{*}{Highway}    & Red & 0.610 & 0.652 & 0.724 & $-0.0113x^3 - 0.0109x^2 + 0.1219x + 0.0848$ \\
                            & NIR & 0.123 & 0.768 & \textbf{0.946} & $-0.0150x^3 - 0.0361x^2 + 0.0562x + 0.1876$ \\
\midrule
\multirow{2}{*}{Industrial} & Red & 0.378 & 0.706 & \textbf{0.994} & $-0.0181x^3 - 0.0245x^2 + 0.1349x + 0.1166$ \\
                            & NIR & 0.592 & 0.901 & \textbf{0.987} & $-0.0029x^3 + 0.0069x^2 + 0.0294x - 0.0455$ \\
\midrule
\multirow{2}{*}{Pasture}    & Red & 0.288 & 0.638 & 0.849 & $-0.0108x^3 - 0.0176x^2 + 0.0812x - 0.0429$ \\
                            & NIR & 0.000 & 0.004 & 0.018 & $0.0026x^3 - 0.0019x^2 - 0.0137x - 0.0224$ \\
\midrule
\multirow{2}{*}{PermanentCrop} & Red & 0.008 & 0.060 & 0.284 & $-0.0155x^3 - 0.0095x^2 + 0.0767x + 0.1168$ \\
                               & NIR & 0.091 & 0.716 & \textbf{0.984} & $-0.0165x^3 + 0.0319x^2 + 0.1057x - 0.0778$ \\
\midrule
\multirow{2}{*}{Residential} & Red & 0.616 & 0.622 & 0.853 & $0.0065x^3 - 0.0013x^2 - 0.0546x + 0.1381$ \\
                             & NIR & 0.012 & 0.313 & \textbf{0.994} & $-0.0254x^3 + 0.0214x^2 + 0.1284x - 0.0506$ \\
\midrule
\multirow{2}{*}{River}      & Red & 0.017 & 0.106 & \textbf{0.991} & $0.0339x^3 + 0.0136x^2 - 0.1891x + 0.0233$ \\
                            & NIR & 0.836 & 0.852 & 0.879 & $-0.0044x^3 + 0.0043x^2 + 0.0698x + 0.0022$ \\
\midrule
\multirow{2}{*}{SeaLake}    & Red & 0.565 & 0.566 & 0.617 & $-0.0055x^3 + 0.0002x^2 + 0.0628x - 0.0739$ \\
                            & NIR & 0.447 & 0.460 & 0.746 & $0.0185x^3 + 0.0049x^2 - 0.1420x - 0.0568$ \\
\bottomrule
\end{tabular}
}
\end{table*}

\subsection{Comparative Case Study and Spectral Reasoning Maps}

\begin{figure*}[htbp]
    \centering
    \includegraphics[width=\textwidth]{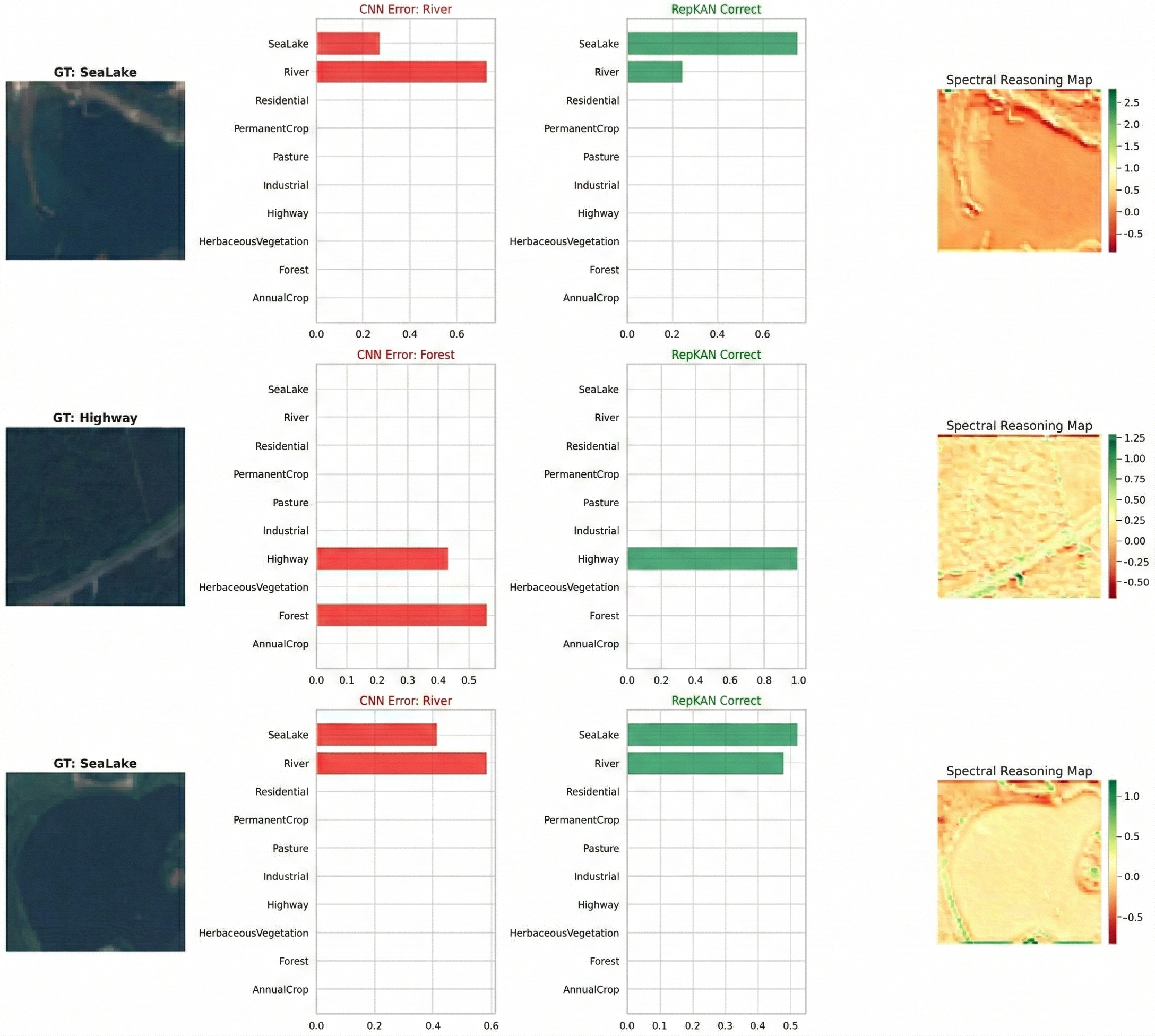}
    \caption{Comparative case study and spectral reasoning visualization. A row-wise comparison between a baseline CNN and RepKAN on selected EuroSAT samples where the CNN fails but RepKAN succeeds. Column 1: Ground-truth (GT) RGB images. Column 2: Class probability distributions from the baseline CNN, highlighting misclassification errors (red bars). Column 3: Correct classification results from RepKAN (green bars). Column 4: Spectral reasoning maps from RepKAN's first stage, illustrating the internal evidence used to distinguish ambiguous land-cover types. The reasoning maps (RdYlGn scale) confirm that RepKAN's success is rooted in its ability to identify class-specific non-linear spectral interactions.}
    \label{fig:comparative_case_study}
\end{figure*}

We conducted a comparative case study against a baseline CNN on challenging EuroSAT samples. As shown in Fig.~\ref{fig:comparative_case_study}, the baseline CNN often struggles with spectrally similar classes, such as SeaLake vs. River and Highway vs. surrounding Forest.

In contrast, RepKAN consistently achieves correct classification by leveraging its Spectral Non-linear Path. The ``Spectral Reasoning Map''---derived from the activation manifolds of Stage 1---reveals that RepKAN focuses on distinct spectral signatures that are invisible to the baseline CNN. For example, in the SeaLake cases, while the CNN is confused by the spatial proximity to river-like banks, the RepKAN reasoning map shows strong positive activations (green regions) specifically tuned to the deep-water spectral characteristics. This granular level of spectral-spatial integration allows RepKAN to serve as a robust, physically interpretable model, effectively correcting the systematic errors found in standard spatial-only networks.

\begin{figure*}[htbp]
    \centering
    \includegraphics[width=\textwidth]{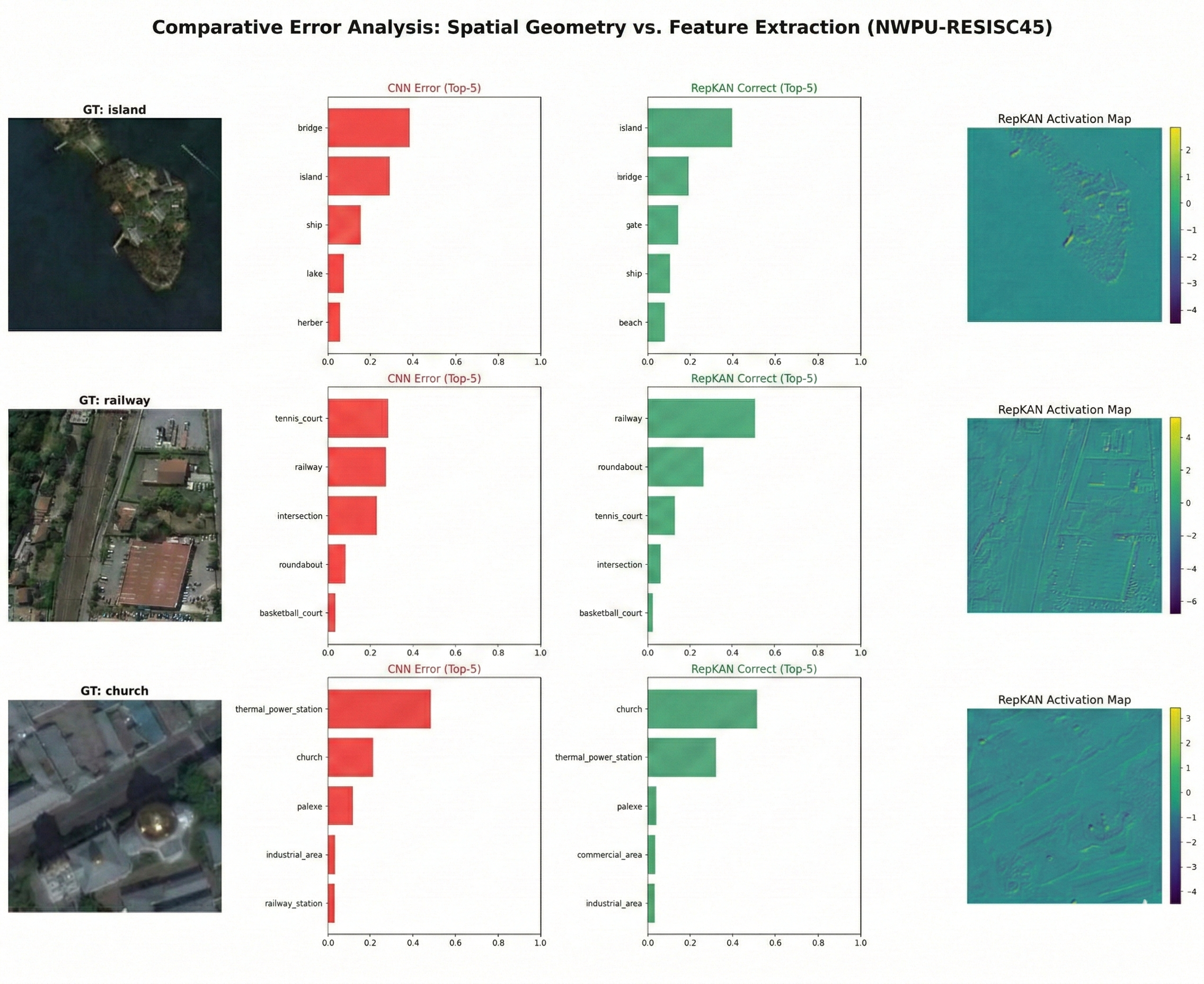}
    \caption{Performance validation on high-resolution aerial imagery (RESISC45). Qualitative comparison between a baseline CNN and RepKAN on semantically complex scenes. Column 1: Ground-truth (GT) aerial images. Column 2: CNN top-5 probability distributions showing failure cases due to structural ambiguity (red bars). Column 3: RepKAN top-5 distributions showing robust and correct classification (green bars). Column 4: RepKAN activation maps illustrating the internal feature extraction process. The maps demonstrate how RepKAN isolates discriminative structural and spectral features to resolve errors typical of standard spatial-only networks.}
    \label{fig:resisc45_case_study}
\end{figure*}

We further conducted a comparative analysis on the NWPU-RESISC45 dataset, which contains high-spatial-resolution aerial imagery with structurally diverse scenes. As illustrated in Fig.~\ref{fig:resisc45_case_study}, baseline CNNs frequently exhibit ``semantic aliasing,'' misidentifying complex land-use categories due to shared textural primitives. For instance, the CNN misclassifies an island as a bridge or ship due to the surrounding water context, and mistakes a church for a thermal power station likely due to similar roof geometries.

RepKAN resolves these ambiguities by integrating non-linear spectral-spatial features. The RepKAN Activation Map (Column 4) provides visual evidence: compared with standard CNN feature maps, it highlights structural boundaries and material-specific signatures that are key to disambiguation. In the railway example, RepKAN isolates linear rail infrastructure despite neighboring industrial textures. These results support RepKAN as an effective framework for high-resolution remote sensing scene understanding.

\section{Discussion and Conclusion}

In this paper, we introduce a novel hybrid architecture for remote sensing image classification, referred to as RepKAN. RepKAN concurrently harnesses the structural advantages of CNNs and the non-linear representational power of KANs. By integrating a Spatial Linear Path with a Spectral Non-linear Path, the model alleviates the ``black-box'' limitations of traditional deep learning architectures. A key contribution of RepKAN is its interpretable design, which discovers class-specific spectral fingerprints and non-linear interaction manifolds related to physical indices such as NDVI. Compared with standard CNNs, RepKAN reduces semantic aliasing in both satellite and high-resolution aerial imagery through spectral reasoning maps. Across EuroSAT and NWPU-RESISC45, RepKAN consistently outperforms the baseline CNN by capturing discriminative spectral-spatial features, indicating its potential as an interpretable backbone for future remote sensing foundation models.

\clearpage
\bibliographystyle{unsrt}
\bibliography{references}

\end{document}